\documentclass[runningheads]{llncs}
\usepackage[T1]{fontenc}
\usepackage{graphicx}
\usepackage{multirow}
\usepackage{hyperref}
\usepackage{comment}
\usepackage{mathtools}
\usepackage{subcaption} 
\usepackage{xcolor}

\begin{document}
\title{ICPR 2024 Competition on Domain Adaptation and GEneralization for Character Classification (DAGECC)}
\titlerunning{DAGECC Competition}
%
\author{Sofia Marino\inst{1}
\and
Jennifer Vandoni \inst{1}
\and
Emanuel Aldea \inst{2}
\and
Ichraq Lemghari \inst{1}\inst{2}
\and
Sylvie Le Hégarat-Mascle \inst{2}
\and
Frédéric Jurie \inst{3}}
\authorrunning{S. Marino et al.}
%
\institute{Safran Tech, Digital Sciences \& Technologies Department, Magny-Les-Hameaux, France 
\and
SATIE CNRS UMR 8029, Paris-Saclay University, Gif-sur-Yvette, France
\\
\and
University of Caen Normandie, France\\
}
\maketitle              
\begin{abstract}
In this companion paper for the DAGECC (Domain Adaptation and GEneralization for Character Classification) competition organized within the frame of the ICPR 2024 conference, we present the general context of the tasks we proposed to the community, we introduce the data that were prepared for the competition and we provide a summary of the results along with a description of the top three winning entries. The competition was centered around domain adaptation and generalization, and our core aim is to foster interest and facilitate advancement on these topics by providing a high-quality, lightweight, real world dataset able to support fast prototyping and validation of novel ideas. 

\end{abstract}
\section{Introduction}
In the rapidly evolving field of computer vision, the ability to develop robust and accurate models that can perform well across diverse scenarios is of paramount importance. Two critical concepts that have emerged as cornerstones in achieving this objective are domain adaptation and domain generalization.

Domain adaptation is a transfer learning technique that aims to improve the performance of models in a target domain by leveraging knowledge from a different but related source domain. This approach is crucial because traditional machine learning models often struggle to generalize well when the data distribution changes between the training (source) and testing (target) domains. In real-world applications, such as autonomous driving \cite{shan2019pixel}\cite{zhang2017curriculum}\cite{li2023domain}, surveillance systems \cite{himeur2023video}\cite{ciampi2023unsupervised}\cite{li2023logical} or industrial quality control \cite{marino2020unsupervised}\cite{thota2020multi}\cite{zhang2021visual}, this shift in data distribution is a common occurrence due to factors like varying weather conditions, lighting, or camera angles. Domain adaptation, therefore, plays a vital role in enhancing the model's resilience and adaptability to these changes, ensuring consistent performance and reliability.

On the other hand, domain generalization is a technique that seeks to develop models that can generalize well to any unseen target domain, without requiring access to any target domain data during training. This is a more challenging task than domain adaptation, as it requires the model to learn a universal representation that can capture the underlying data structure across multiple domains. The importance of domain generalization lies in its potential to create models that can truly adapt to any new environment, making them highly valuable for real-world applications where the target domain may be unknown or constantly changing \cite{choi2021robustnet}\cite{liu2023adversarial}\cite{chen2024deep}\cite{hemadou2024beyond}.

The importance of datasets in the context of domain adaptation and domain generalization cannot be overstated. Diverse and representative datasets are crucial for training models that can effectively learn and adapt to different domains. For domain adaptation, the availability of well-labeled source and labeled or unlabeled target domain datasets is essential, as it allows the model to learn the underlying data distributions and map the knowledge from the source to the target domain. On the other hand, domain generalization typically relies on multi-domain datasets, where the model is trained on data from several different but related domains. This exposure to a variety of data distributions during training helps the model to learn a more universal representation, enabling it to generalize better to unseen target domains. Moreover, the quality and diversity of datasets can significantly influence the performance of these techniques, making the curation and selection of appropriate datasets a critical aspect of their success. Thus, the development and availability of comprehensive and diverse datasets are vital for advancing research and applications in domain adaptation and domain generalization.

The competition on Domain Adaptation and GEneralization for Character Classification (DAGECC) aims to foster advancements in domain adaptation and domain generalization techniques by introducing a novel dataset comprising real images of digits and characters captured from serial numbers on manufactured objects. 
In particular, we focus on an image classification task for industrial serial number recognition. 
In industrial environments, serial part numbers play a critical role in ensuring traceability and streamlining the management of diverse components (e.g., preventive maintenance, analytics, etc.). However, the manual process of reading and recording these serial numbers is both labor-intensive and prone to errors. As a result, the development of an automated serial number recognition system that can work seamlessly across multiple parts holds tremendous potential for improving operational efficiency in industrial environments.

To address this challenge, we are introducing Safran-MNIST, a brand new dataset suite that comprises images of serial numbers extracted from diverse avionic parts manufactured by Safran, an international high-technology group and world leader operating in the aviation (propulsion, equipment and interiors), defense and space markets. The content resembles that of the well-known MNIST dataset of handwritten digits \cite{lecun1998gradient}, hence the name, but with a focus on industrial contexts. The Safran-MNIST dataset suite offers a realistic representation of industrial serial number images, encompassing variations in lighting conditions, orientations, writing styles and surface textures. Thus, it represents a good compromise between 1) requiring a low computational cost for processing, in line with other MNIST variants \cite{lecun1998gradient}\cite{DBLP:conf/icml/GaninL15}, and 2) being highly relevant for real-world applications.

\section{Competition overview}
The competition encompasses two primary tasks: (i) Domain Generalization and (ii) Unsupervised Domain Adaptation. The central objective of this competition is to investigate and foster progress in domain adaptation and domain generalization methodologies, specifically in the context of character recognition. 
\subsection{Datasets}
DAGECC competition is based on the brand new Safran-MNIST dataset suite, specially crafted for the application of character recognition.
Characters have been extracted from larger images of Safran's avionic parts acquired from aircraft engines returned from flights, that have to be periodically inspected for predictive maintenance and security reasons. They correspond to product references or serial numbers engraved on metal parts with various techniques which are widely used for traceability (laser, pencil, or micro percussion engraving).

\begin{figure}[h]
    \includegraphics[width=\textwidth]{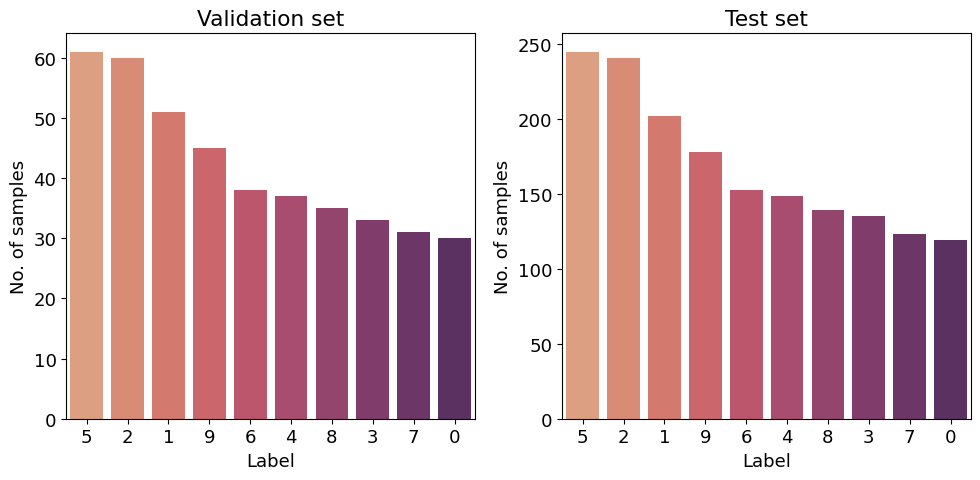}
    \caption{Distribution of class samples in Safran-MNIST-D by sets.}
    \label{fig:image4}
\end{figure}

\begin{figure}
    \centering
    \begin{subfigure}[b]{0.48\textwidth}
        \includegraphics[width=\textwidth]{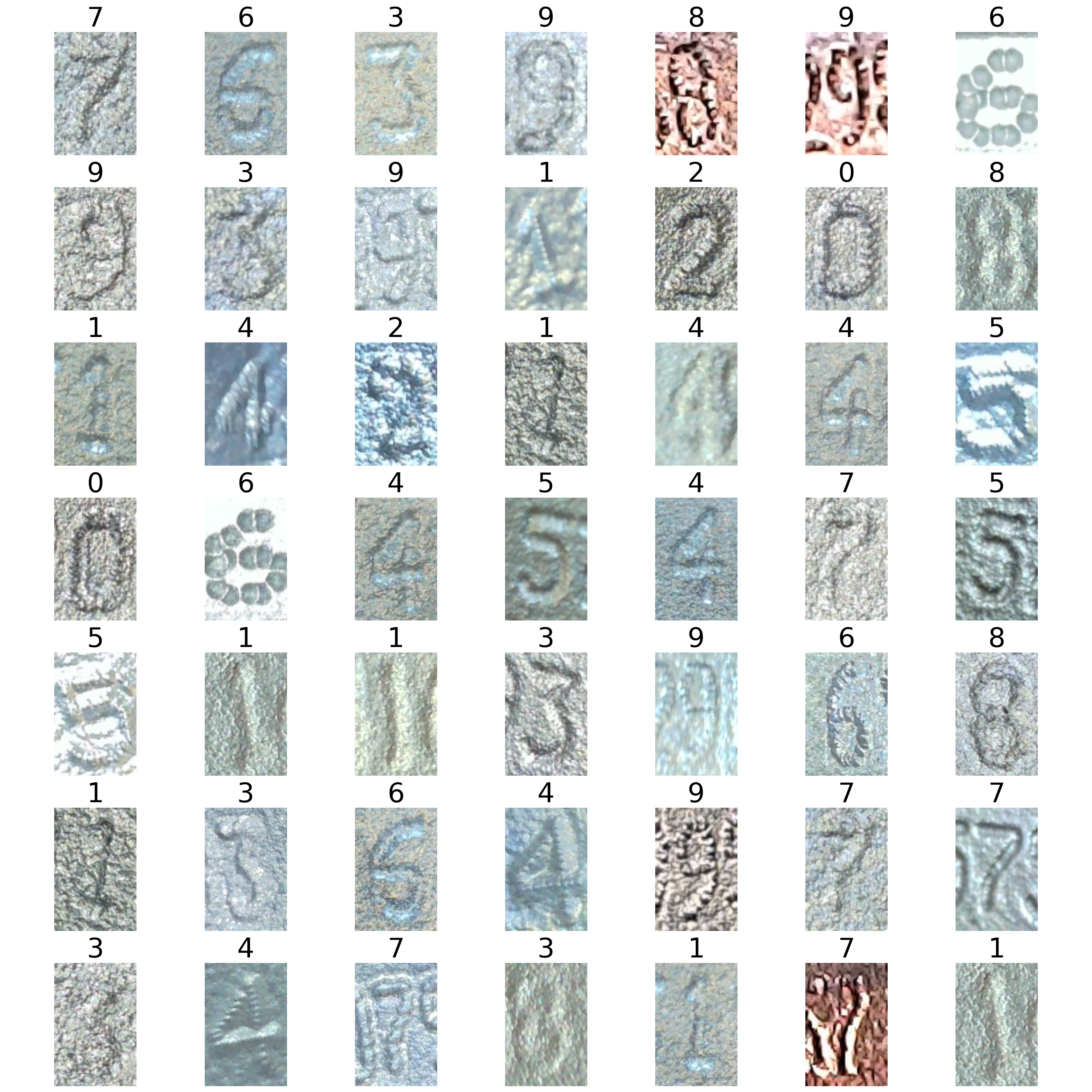}
        \caption{Safran-MNIST-D images.}
        \label{fig:image1}
    \end{subfigure}
    \hfill
    \begin{subfigure}[b]{0.48\textwidth}
        \includegraphics[width=\textwidth]{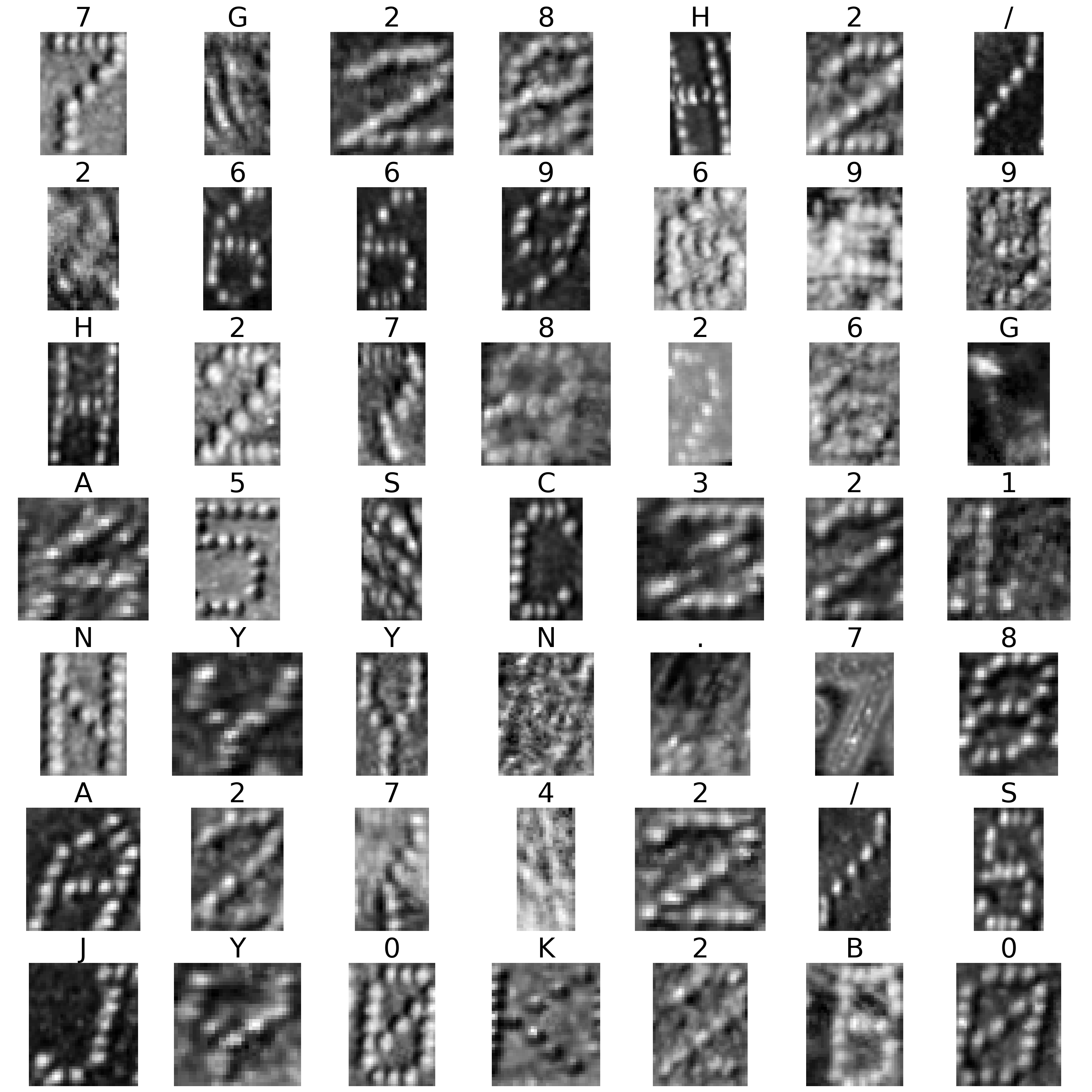}
        \caption{Safran-MNIST-DLS images.}
        \label{fig:image2}
    \end{subfigure}
    \caption{Images extracted from Safran-MNIST dataset suite.}
    \label{fig:both_images}
\end{figure}

The Safran-MNIST dataset suite comprises two datasets, namely Safran-MNIST-D\footnote{\url{https://zenodo.org/records/13320997}}\cite{databaseD} and Safran-MNIST-DLS\footnote{\url{https://zenodo.org/records/11093441}}\cite{databaseDLS}, which are used in the two tasks respectively. Specifically: 
\\ \\
\textbf{Dataset for Task 1 - Domain Generalization}:
Target data for this task consists of real-world Safran-MNIST-D dataset containing RGB images of size $128 \times 192$ pixels representing digits ranging from 0 to 9. Each image in this dataset belongs to a distinct class. We annotated 1684 images for testing and 421 images for validation. Figure \ref{fig:image4} shows the class distribution for each set. Since target data is not to be used in this task, the validation and testing datasets and associated ground-truth are released at the end of the competition. Sample examples from this dataset are shown in Figure \ref{fig:image1}.  

\begin{figure}[htbp]
    \includegraphics[width=\textwidth]{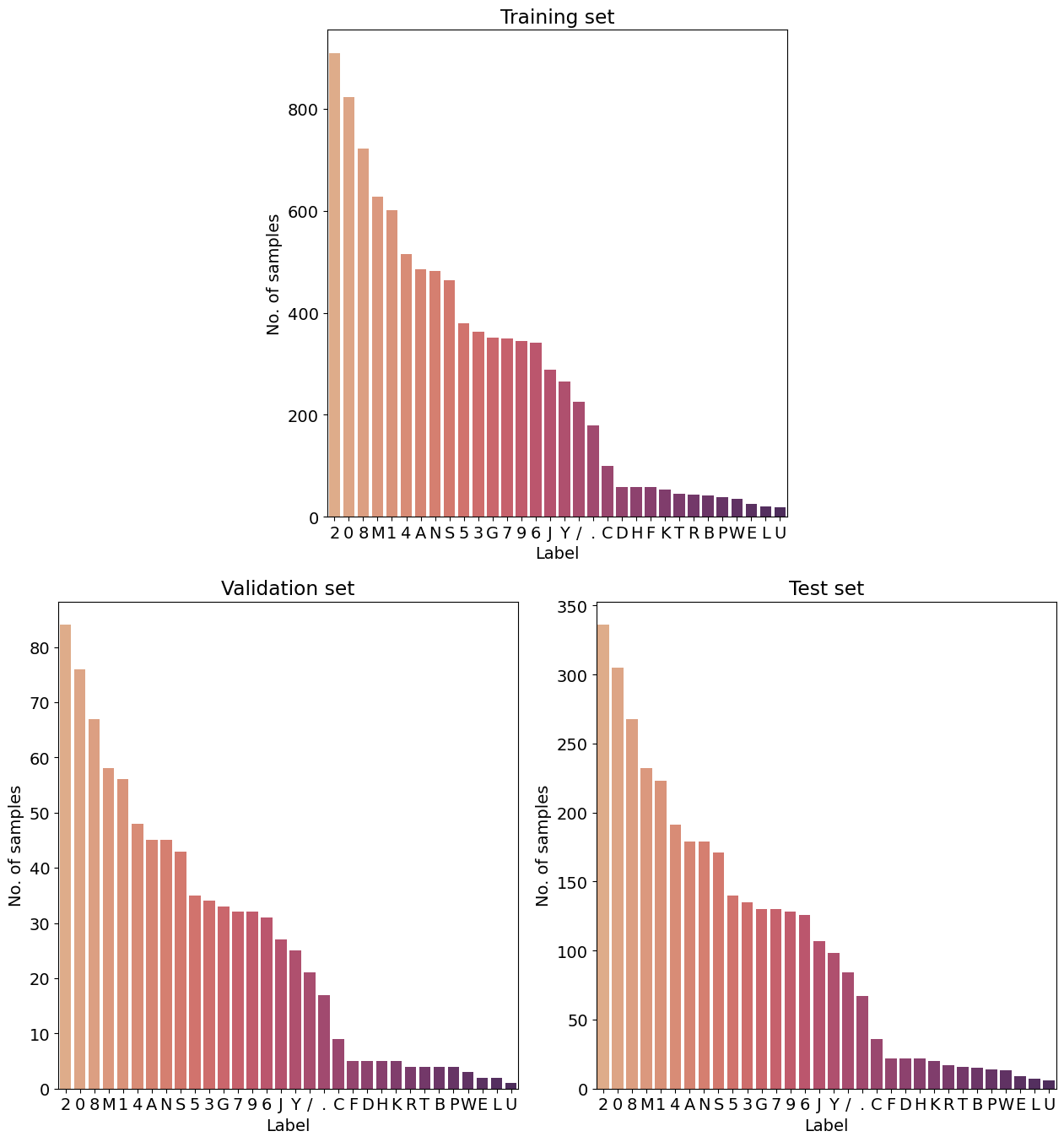}
    \caption{Distribution of class samples in Safran-MNIST-DLS by sets.}
    \label{fig:image5}
\end{figure}

\noindent \textbf{Dataset for Task 2 - Unsupervised Domain Adaptation}:
Target data for this task consists of real-world Safran-MNIST-DLS dataset containing variable size\footnote{from $18 \times 30$ pixels to $86 \times 79$ pixels} gray-scale images of 32 characters. Specifically, this dataset includes 10 digits ranging from 0 to 9, 20 letters (A, B, C, D, E, F, G, H, J, K, L, M, N, P, R, S, T, U, W and Y) and 2 symbols (/ and .). Each image in this dataset belongs to a distinct class. We manually annotated 3448 images for testing and 862 for validation. Furthermore, we provided 9314 unlabeled images for training. Figure \ref{fig:image5} shows the class distribution for each set. The validation and testing datasets are released at the end of the competition as well as the ground-truth associated to all of the sets. Sample examples from this dataset are  shown in Figure \ref{fig:image2}. 
\\ \\ 
An overview of both datasets is shown in Table \ref{tab:overview}.
\begin{table}[htbp]
\setlength\tabcolsep{8pt} 
\centering
\caption{Overview of Safran-MNIST-D and Safran-MNIST-DLS datasets.}
\label{tab:overview}
\begin{tabular}{lccccc}
\hline
\multirow{2}{*}{Dataset} & \multicolumn{3}{c}{Number of Images} & \multirow{2}{*}{Data Type} &\multirow{2}{*}{\begin{tabular}{@{}c@{}}Number\\ of Classes\end{tabular}} \\ \cline{2-4}
                         & Training  & Validation  & Test    &            &                  \\ \hline
Safran-MNIST-D           &     -       &     421   &  1684   & RGB & 10                 \\ 
Safran-MNIST-DLS         &      9314      &        862      &    3448     &   Gray-scale & 32                  \\ \hline
\end{tabular}
\end{table}
\setlength\tabcolsep{6pt} 
\subsection{Description of tasks}
\subsubsection{Task 1 - Domain Generalization:}
The aim of this task is to develop models that can
generalize well to an unseen target domain, without requiring access to any target domain data during
training. Thus, participants could not use any data from the target domain, which as the new Safran-MNIST-D dataset that contains images of numbers ranging from 0 to 9. In addition, participants were granted the freedom to use publicly available data or generated data from various source domains, such as MNIST \cite{lecun1998gradient}, MNIST-M \cite{ganin2016domain}, SVHN \cite{37648}, HASYv2 \cite{thoma2017hasyv2}, DIGITS \cite{alpaydin1998optical}, EMNIST \cite{cohen2017emnist}. This allowed participants to leverage existing datasets or generate realistic synthetic data to train their models without compromising the requirement of no access to the target domain. Note that we deliberately not imposed any dataset for the source domain in order to let participants find and explore relevant  datasets (or combinations/derivations of datasets) adapted to the task, provided that these are \emph{publicly available}. Proprietary data were not allowed in order to facilitate the reproducibility of results. 

\subsubsection{Task 2 - Unsupervised Domain Adaptation:}
This task is focused on unsupervised domain adaptation methods, in which we provided unlabeled data from a target domain: the new Safran-MNIST-DLS dataset, which comprises images of 32 classes depicting numbers, alphabetic characters, and symbols. Participants had access to the unlabeled target data during training. Furthermore, participants were tasked with sourcing a suitable dataset that can serve as the \textit{source data} for domain adaptation. This could involve either finding an existing publicly available dataset or generating a synthetic dataset that aligns with the problem domain, using traditional image processing techniques or generative AI provided that the generative model has been trained on public data. \\ \\
For both tasks, participants could use any publicly available and appropriately licensed data to pre-train their models. 

Note that data from one task of the competition could not be exploited for the other task.

\subsection{Evaluation metric}
Macro-averaged F$_1$-score was selected as evaluation metric in this competition to take into account the imbalanced
nature of the dataset. This metric is described as follows:
\begin{equation}
F_1^{Macro}=\frac{\sum_{k=1}^KF_1^k}{K}.
\end{equation}
For class $k$,
\begin{equation}
F_1^k=\frac{2TP^k}{2TP^k+FP^k+FN^k},
\end{equation}
where \(\mathit{TP^k}\) is the number of true positives for class \(\mathit{k} \), \(\mathit{FP^k}\) is false positives of class \(\mathit{k}\) and \(\mathit{FN^k}\) is false negatives of class \(\mathit{k}\).
\subsection{Submisison system}
Codabench \cite{codabench} was used for submissions. Codabench is an open source platform that efficiently facilitates the organization of competitions. Participants could try out their methods, get real-time feedback and results on a competitive leaderboard.
Each task had its individual competition website where participants had the opportunity to make submissions during two distinct phases. The first of these was the development phase, where the models were evaluated on the unseen validation set. In this phase, each teams had a quota of six submissions per day. This was followed by the final phase, during which the best final model selected by the teams was evaluated only once on the unseen test set.

\section{Competition Results and Analysis}
\begin{figure}[htbp]
    \includegraphics[width=\textwidth]{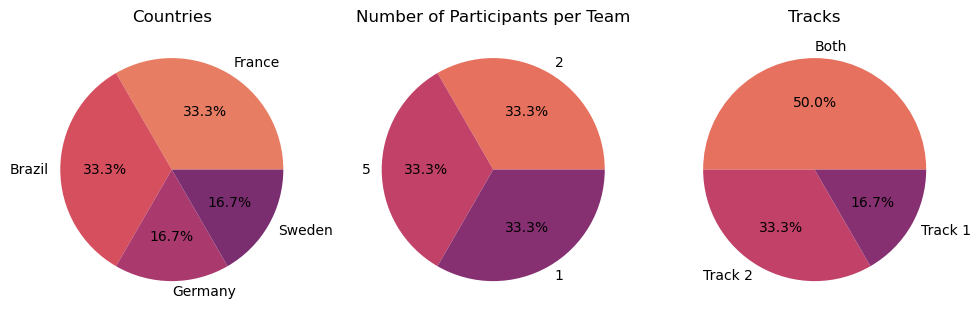}
    \caption{Breakdown of competition participation.}
    \label{fig:image3}
\end{figure}

A total of 28 teams registered for the competition using the competition website\footnote{\url{https://dagecc-challenge.github.io/icpr2024/}} and six of them made submissions on Codabench (including for both tasks). A total of 125 submissions were recorded for Task 1, compared to 181 for Task 2, indicating strong participant activity. Figure \ref{fig:image3} summarizes the information of active teams. 
\subsection{Task 1: Domain Generalization}
Results for Task 1 are shown in Table \ref{tab:task1}. Here we briefly describe the top three winners of this task. 
\begin{enumerate}
    \item \textit{Team Deng}: constituted of two people, Victor Deng, student from ENS Paris in France, and Erwin Deng, student from CentraleSupelec also situated in France. The winning solution relies on the ResNet50 \cite{he2016deep} architecture, initialized with pre-trained weights on the ImageNet \cite{deng2009imagenet} dataset. Several specific training techniques were used. The model was fine-tuned using a custom synthetic dataset, along with MNIST, SVHN, and MNIST-M. Due to the imbalanced nature of the dataset, the team used the WeightedRandomSampler provided by PyTorch \cite{Ansel_PyTorch_2_Faster_2024} during training. To generate the custom dataset, they proposed the following method. Firstly, the background was generated using a random color, a random noise, followed by an embossing effect with custom filters, and a Gaussian blur, mimicking real-world imperfections. Secondly, many open-source fonts were used to generate digits from 0 to 9. Additionally, geometric transformations were applied to these digits (random rotations, scaling, translations). Finally, digits and backgrounds were combined with a random intensity followed by embossings effects, blurring and more.
    \item \textit{Fraunhofer IIS DEAL}: Anne Klier, Sai Rahul Kaminwar, Gabriel Dax, Sina Emami and Mohamed Hesham Ibrahim Abdalla from Fraunhofer IIS DEAL in Germany. The team used a GoogLeNet~\cite{szegedy2015going} pretrained on ImageNet and fine-tuned on various public datasets (MNIST, Chars74K, HASYv2, SVHN, USPS, SYN NUMBERS, MNIST-M). In addition, a synthetic dataset of 500 images per digit was generated using a pretrained stable diffusion model \cite{esser2024scaling} which was required to provide weathered images of digits embossed or engraved on a metallic surface.
    \item \textit{JasonMendoza2008}: Romain Lhotte, registered as an individual participant, is a student from CentraleSupelec, located in France. His solution is based on gathering as much data from publicly available datasets, until reaching 200,000 images. Then, prediction was achieved by a weighted mean of the predictions of five distinct networks. These networks are based on AlexNet \cite{krizhevsky2012imagenet} and a baseline CNN architecture that was provided by organizers as example. The latter contains two convolutional layers: the first layer containing 16 channels with a kernel size of 5, and a second layer containing 32 channels with a kernel size of 5. Each convolutional layer is followed by a rectified linear unit (ReLU) activation function along with a MaxPooling layer.
\end{enumerate}
Analyzing the top-performing solutions for Task 1 - Domain Generalization reveals common strategies among the participating teams that likely contributed to their success. 
Each team leveraged pretrained deep learning architectures, such as ResNet50, GoogLeNet, and AlexNet, initialized with ImageNet weights. In an interesting differentiation strategy, the top two teams turned to synthetic datasets. Each of them crafted unique data generation methods: the winner utilized conventional image processing techniques, while the runner-up chose to leverage generative AI methods. Underscoring the importance of vast and diverse datasets, all teams focused on amassing as much data as possible from various sources. 
Lastly, only the winning team deployed a technique to mitigate the challenge of imbalanced datasets. 

\begin{table}[h]
\centering
\caption{Competition results (list of all successful submissions on the final phase) for Task 1: Domain Generalization.}
\begin{tabular}{c|c|c|c}
Rank & Team name & Codabench username & Macro F1-score \\
\hline
1 & Team Deng & eternel22 & 0.82 \\
2 & Fraunhofer IIS DEAL & Fraunhofer IIS DEAL & 0.74 \\
3 & JasonMendoza2008 & JasonMendoza2008 & 0.50 \\
4 & Deep Unsupervised Trouble & heitor & 0.44 \\
\hline

\end{tabular}
\label{tab:task1}
\end{table}

\subsection{Task 2: Domain Adaptation}
The results for Task 2 are shown in Table \ref{tab:task2}. 
We describe briefly the top three winners of this task: 
\begin{enumerate}
    \item \textit{Team Deng}: this is the same team that won Task 1 of competition. The proposed winning solution for this track bears similarity to the solution suggested for Track 1. Specifically, the pre-trained ResNet50 was employed as the architecture for classifying 32 different classes, including digits, letters, and symbols. The distinguishing factor lies in the datasets used for this track, which include EMNIST and custom-generated data. The procedure for generating this latter dataset closely aligns with the approach used in Track 1.
    Available unlabeled Safran-MNIST-DLS data was not used.
    \item \textit{Deep Unsupervised Trouble}: Artur Jordão Lima Correia, Bruno Lopes Yamamoto, Heitor Gama Ribeiro, Leandro Giusti Mugnaini, Victor Sasaki Venzel from the Escola Politécnica of the University of São Paulo. EMNIST~\cite{cohen2017emnist} was used as the primary training dataset, along with additional samples generated using the Consola font in order to mimic the dotted style of the original data. Then, the competitors applied a preprocessing step based on Otsu thresholding~\cite{otsu1975threshold}, which was used to increase the training data size, as follows. From a single input image, multiple character versions were created by using binarization thresholds varying around the optimal value provided by Otsu's algorithm, and then entropy and contrast filters were applied. A standard ResNet18~\cite{he2016deep} architecture was used with the augmented dataset.
    \item \textit{Raul}: Raul Cavalcante, from Instituto de Matemática e Estatística da Universidade de São Paulo. The competing team generated a large amount of synthetic data (approximately 240k images) by creating 3D-rendered characters engraved into metallic plates using Blender \cite{blender}. Data diversity was obtained by varying programmatically parameters such as material properties, camera angles, deletion of random dots and applying random noise to dot locations. Then, a CNN architecture with residual connections was used for training.
\end{enumerate}
In analyzing the methods employed by the top three teams for Task 2: Domain Adaptation, several prevailing trends emerge. Firstly, the top two teams leveraged pretrained deep learning architectures, such as ResNet50 and ResNet18, initialized with ImageNet weights. Secondly, it is interesting to note that none of the top three teams took advantage of the Safran-MNIST-DLS unlabeled dataset. Instead, the participating teams opted to generate their own datasets or utilize widely recognized datasets such as EMNIST. \\
Although the best performing methods relied on synthetic data generation, we commend the effort of the Machine Learning Group LTU team (Simon Corbillé and Elisa H. Barney Smith, Lule\aa\ University of Technology, Sweden) which decided to focus more than the competitors on the unlabeled corpus. The team proposed a learning process that trains a ResNet-34 model with labeled data, then cycles through a 1) prediction and pseudo label generation on the unlabeled data, 2) majority voting and 3) retraining using real labels and pseudo labels in order to maintain stability. We argue that, beyond the low hanging fruit of data generation, the most promising perspective for further improving the task performance is to better exploit the information contained in the unlabeled corpus. 
\begin{table}[h]
\centering
\caption{Competition results (list of all successful submissions on the final phase) for Task 2: Domain Adaptation.}
\begin{tabular}{c|c|c|c}
Rank & Team name & Codabench username & Macro F1-score \\
\hline
1 & Team Deng & eternel22 & 0.65 \\
2 & Deep Unsupervised Trouble & heitor & 0.6 \\
3 & Raul & raulcd & 0.52 \\
4 & Machine Learning Group LTU & simcor & 0.44 \\
5 & JasonMendoza2008 & JasonMendoza2008 & 0.14 \\
\hline
\end{tabular}
\label{tab:task2}
\end{table}

\section{Conclusion}
In this paper, we have presented the competition on Domain Adaptation and GEneralization for Character Recognition. We have introduced a new real-world dataset suite, Safran-MNIST, which comprises two datasets, Safran-MNIST-D and Safran-MNIST-DLS. A total of six teams submitted results on two different tasks, namely Domain Adaptation and Domain Generalization. The winner of both tracks was Team Deng from CentraleSupelec and ENS Paris, both institutions located in France. As expected given the lightweight format of the data, the architectures used by the participants were relatively compact, and their efforts focused mainly on scraping and generating synthetic data samples akin to the target distribution. We hope that the interest generated by the competition and the ease of use of the proposed datasets will sustain the improvement efforts in the future, and benefit the worldwide research community and the industry. 
\\ \\

\subsubsection{Acknowledgements} A special thank to SAFRAN, especially Basile Musquer (SAFRAN Aircraft Engines) and Thierry Arsaut (SAFRAN Helicopter Engines) for participating in the acquisition of the images, the creation of the dataset and for allowing the public release of the data. We are also grateful to Codabench, and specifically to Adrien Pav\~ao, for their great support. 

%
%
%
%
\bibliographystyle{splncs04}  
\bibliography{bib}






\end{document}